\documentclass[letterpaper, 10pt, conference]{ieeeconf}
\IEEEoverridecommandlockouts
\usepackage{graphicx} %
\usepackage{balance}

\usepackage{amsmath, amssymb}   %
\usepackage{threeparttable}     %
\usepackage{multirow}           %
\usepackage{booktabs}           %
\usepackage{xcolor}
\usepackage{tabularx}
\usepackage{fontawesome5}

\definecolor{mylinkcolor}{RGB}{40, 115, 201}
\definecolor{mycitecolor}{RGB}{71, 191, 38}
\definecolor{DGreen}{RGB}{107, 190, 35}
\definecolor{mygreen}{RGB}{106, 168, 79}
\definecolor{myblue}{RGB}{60, 120, 216}
\definecolor{myred}{RGB}{204, 0, 0}

\newif\ifshowchanges
\showchangestrue          %

\ifshowchanges

  \newcommand{\deleted}[1]{\textcolor{rbModified}{\sout{#1}}}
\else

  \newcommand{\deleted}[1]{}   %
  
\fi

\makeatletter
\let\NAT@parse\undefined
\makeatother
\usepackage[bookmarks=true, colorlinks=true, citecolor=mycitecolor,linkcolor=mylinkcolor,urlcolor=mycitecolor]{hyperref}
\usepackage[noabbrev,capitalise]{cleveref}

\usepackage{makecell}
\usepackage{xspace}
\usepackage{stfloats}
\usepackage{caption}
\usepackage[normalem]{ulem}                 %
\definecolor{rbAdded}{HTML}{0B5FA5}
\definecolor{rbModified}{HTML}{B00020}
\definecolor{rbDraft}{HTML}{7B3FA0}
\newcommand{\jaxipmnoIL}{\mbox{\texttt{jaxipm}-noIL}\xspace}  %

\title{\LARGE \bf Scaling Nonlinear Optimization: Many Problems One GPU}

\author{
John Viljoen$^{1}$,
Johanna Haffner$^{2}$,
Lasse Peters$^{1}$
Masayoshi Tomizuka$^{1}$, and
Negar Mehr$^{1}$
\thanks{$^{1}$J. Viljoen, M. Tomizuka, and N. Mehr are with the Department of Mechanical Engineering, University of California, Berkeley, CA, USA.
        {\tt\small \{john\_viljoen, lasse.peters, tomizuka, negar\}@berkeley.edu}}
\thanks{$^{2}$J. Haffner was with the Department of Biosystems Science and Engineering, ETH Z\"urich, Switzerland, during the completion of this work.
        {\tt\small johanna@haffner.dev}},
\thanks{This work has been submitted to the IEEE for possible publication. Copyright may be transferred without notice, after which this version may no longer be accessible.}
}

\begin{document}
\typeout{TOPMARGIN=\the\topmargin}
\typeout{HEADHEIGHT=\the\headheight}
\typeout{HEADSEP=\the\headsep}
\typeout{TEXTHEIGHT=\the\textheight}
\maketitle

\begin{figure*}[!ht]
	\centering
	\includegraphics[width=0.70\textwidth]{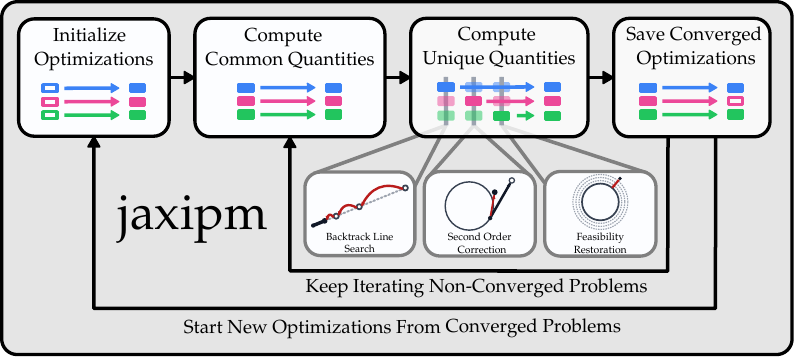}
	\caption{\small We summarize \texttt{jaxipm}: each color represents a different optimization problem being executed in batch. At each iteration, every problem can take different paths, including but not limited to: backtracking line search, second-order correction, and feasibility restoration. Each of these heterogeneous paths shares some common quantities in their calculation, which we compute in parallel, followed by the remaining unique quantities sequentially.}
	\label{fig:teaser}
	\vspace{-1.5em}
\end{figure*}

\begin{abstract}
Modern GPUs and software frameworks have made sampling-based methods, such as reinforcement learning and model predictive path integral control, increasingly popular in robotics. However, the logic these methods can execute per sample is limited to what has been implemented on the GPU.
Nonlinear programs (NLPs) are one of the most expressive tools in robotics, reliably minimizing a cost while exactly enforcing general constraints.
NLP optimizers were among the field's main workhorses, but were not designed for modern GPU hardware.
Embedding NLPs such as trajectory optimization and inverse kinematics directly within each sample would be highly valuable, yet the only current option is to call a CPU solver per sample, possibly in parallel on a multi-core CPU.
This bottlenecks the CPU, requires GPU-to-CPU-to-GPU transfers when the rest of the pipeline resides on the GPU, and leaves the GPU intermittently idle.
To address this issue, in this work we present \texttt{jaxipm}, the first GPU-batched solver for generally constrained nonlinear optimization problems.
Our approach is a JAX implementation of the battle-tested IPOPT algorithm, facilitated by eliminating control flow through \textit{heterogeneous iteration fusion}, and minimizing GPU idle time through \textit{iteration-level batching}.
Our implementation enables solving \emph{hundreds} of NLPs in parallel on a \emph{single} GPU.
We validate \texttt{jaxipm} on a range of quadrotor nonlinear model predictive control benchmarks, where on a single GPU we achieve between $30.7\%$ and $69.9\%$ of the throughput of 64 IPOPT instances running in parallel on a 64-core CPU. 

\smallskip\noindent\href{https://github.com/johnviljoen/jaxipm} {\textcolor{blue!60!black}{\faGithub\enspace{open-source codebase}}}%

\end{abstract}

\vspace{-0em}
\section{Introduction}
Modern GPU-batched simulators~\cite{mjx, isaac_lab, brax} have made sampling-based (SB) methods, such as reinforcement learning (RL), model predictive path integral control (MPPI)~\cite{mppi}, and the cross-entropy method~\cite{cem}, increasingly popular in robotics.
By evaluating many samples concurrently, these frameworks achieve orders-of-magnitude higher sample throughput than classical CPU-based alternatives.
However, the logic that SB methods can invoke per sample is limited to algorithms available on the GPU.
Today, these include neural networks~\cite{pytorch}, physics simulation~\cite{mjx,isaac_lab,brax}, and batched optimizers for convex problems such as linear and quadratic programs~\cite{optnet, qpax, moreau, mpax, jaxopt}, but no solver for general Nonlinear programs (NLPs).

NLPs with general constraints are one of the most expressive tools in robotics. Many core problems, such as trajectory optimization and inverse kinematics, reduce to NLPs \cite{traj_opt_nlp, nmpc_nlp, ik_nlp}.
Mature gradient-based solvers such as IPOPT \cite{ipopt}, KNITRO \cite{knitro}, and SNOPT \cite{snopt} reliably solve such problems.
These methods converge to solutions satisfying optimality conditions, scale well with problem dimension, and exploit model information to refine solutions to high precision.
These desirable features have made NLPs one of the main workhorses of robotics~\cite{nlp_survey}.
Embedding NLPs, such as trajectory optimization and inverse kinematics, directly within each sample would enable dataset-scale collections of constrained, locally optimal solutions.

However, general NLP solvers were never created for the GPU, and therefore their integration in modern SB approaches remains limited.
If a practitioner wanted to integrate a general NLP into their sampling-based approach today, the only option to solve batches of NLPs is to call a CPU solver per sample, potentially in parallel on a multi-core CPU.
When the rest of the pipeline runs on the GPU, however, using a CPU-based NLP solver requires transferring data from the GPU to the CPU and back.
Moreover, combining CPU- and GPU-intensive workloads can leave either device idle while the other operates at full capacity.

In this paper, we introduce \texttt{jaxipm}, the first open-source, GPU-batched solver for generally constrained nonlinear optimization problems, i.e., NLPs with general equality and inequality constraints, capable of solving hundreds of problems concurrently on a single GPU.
We base \texttt{jaxipm} on the battle-tested algorithm of IPOPT~\cite{ipopt}, one of the most widely used NLP optimizers~\cite{casadi, drake, pyomo, jump} due to its broad applicability and robustness.
IPOPT solves an NLP iteratively, and each iteration may resort to a variety of fallbacks for known edge cases.
This machinery underlies IPOPT's robustness but poses unique challenges for GPU-based execution.
Within a batch, every problem can take a different control-flow branch at each iteration and terminate at a different iteration count.
This conflicts with the uniform execution model that GPUs rely on for parallelism.
We get around this issue in \texttt{jaxipm} with two main contributions.

Our first main contribution~\textbf{(C1)} is \textit{heterogeneous iteration fusion}, where we rework every possible fallback in the IPOPT algorithm into one iteration type without losing the benefits of those fallbacks.
Our key insight is that most of IPOPT's control-flow branches \textit{share a common set of expensive computations}, meaning that we can compute these shared computations concurrently across an entire batch.
For the small remainder of truly unique computations, we run them for all problems regardless of the branch IPOPT would have taken, and select the correct output afterwards.

Our second main contribution~\textbf{(C2)} is \textit{iteration-level batching}, where we replace finished optimization problems without interrupting ongoing ones to minimize GPU downtime.
A naive parallel implementation would batch at the solve level, requiring every problem to complete before the next batch can begin and leaving the GPU idle until the slowest solve finishes.
Iteration-level batching minimizes this idling by synchronizing at iteration boundaries and letting completed problems be replaced with new ones mid-batch while unfinished problems continue iterating.

To achieve high performance and to allow for native interoperability with GPU-batched, we implement~\texttt{jaxipm} in JAX \cite{jax} learning frameworks used in modern robotics research \cite{mjx, isaac_lab, brax}.
We evaluate \texttt{jaxipm} on three quadrotor nonlinear model predictive control (NMPC) benchmarks: single-quadrotor navigation, reference tracking with time-varying obstacles, and multi-quadrotor consensus.
On a single GPU, \texttt{jaxipm} achieves between $30.7\%$ and $69.9\%$ of the throughput of 64 IPOPT instances running in parallel on a 64-core CPU. %
We further show that \texttt{jaxipm}’s iterates match IPOPT’s to within $10^{-8}$ at every iteration on a problem that exercises most of IPOPT’s control-flow branches.

\section{Related Works}

\noindent \textbf{NLP Optimizers.} Decades of work have produced mature NLP optimizers that implement sequential quadratic programming \cite{snopt,knitro} and interior-point methods \cite{ipopt,knitro}.
These solvers handle general equality and inequality constrained problems, scale to thousands of variables, and provide convergence guarantees that have made them the backbone of model-based robotics and engineering optimization \cite{casadi,drake,pyomo,jump,bhatt2025strategic}.
However, these methods all target the CPU and execute their iterations sequentially.

\noindent \textbf{GPU Acceleration of NLP Optimizers.} A recent line of work now targets IPOPT-like optimizers by porting their dominant per-iteration cost onto the GPU. These are: the solution of a highly sparse symmetric indefinite KKT system, and the sparse derivative evaluation that constructs that linear system.
HyKKT \cite{hykkt} was the first to apply a positive-definite reformulation of the indefinite KKT system as a GPU-acceleration strategy for general NLPs.
MadNLP \cite{madnlp} reimplements IPOPT in Julia and achieves GPU acceleration in two ways.
Firstly, it supports its own positive-definite KKT system reformulation \cite{lifted_kkt} solved by the GPU-accelerated linear solver cuDSS \cite{cudss}.
Secondly, MadNLP accelerates the evaluation of sparse derivatives through ExaModels \cite{madnlp}.
Despite achieving GPU acceleration, MadNLP does not achieve GPU-\textit{batching} as its algorithmic logic, including its control-flow branches, remains CPU-bound.

\noindent \textbf{Batched Optimizers.} Prior GPU-batched optimizers have primarily focused on convex problems, which require less of the control-flow branching that kept NLP optimizers CPU-bound.
A number of approaches have been used for these, such as interior-point methods \cite{optnet, qpax, moreau}, primal-dual hybrid-gradient methods \cite{mpax}, and operator-splitting methods \cite{jaxopt}.
One line of work has created a trust-region Newton method for GPU-batching of problems with nonconvex objectives \cite{exatron}, but each problem may only have up to~32~optimization variables and only possess bound constraints.
This leaves \emph{general} NLPs arising from applications with nonlinear dynamics or constraints without a batched GPU optimizer.

\noindent \textbf{Demand For Batched NLP Data Generation.}
Recent robotics research reveals a growing demand for large datasets of high-quality optimization solutions.
Opt2Skill~\cite{opt2skill} trains humanoid RL policies on reference trajectories that an optimizer generates offline, showing that model-based optimization produces higher-quality training data than human demonstrations or inverse kinematics baselines.
In~\cite{quadruped_mpc_imitate}, the authors take the same approach for legged locomotion, distilling an MPC controller into a quadruped policy via behavior cloning, where every demonstration in the training set is an NLP solution.
In~\cite{good_old_fashioned_engineering}, the authors argue more broadly that closing the robot learning data gap requires hybrid pipelines combining model-free learning with model-based optimization at scale.
These works point to the same bottleneck: no current optimizer produces NLP solutions on the GPU at the scale these pipelines demand.
We aim to close this gap with \texttt{jaxipm}, which runs \emph{hundreds} of solves concurrently on a GPU.

\section{Problem Statement}

\begin{table*}[t]
	\centering

	\renewcommand{\arraystretch}{1.0}
	\begin{tabularx}{\textwidth}{@{}p{0.24\textwidth}XX@{}}
		\toprule
		\textbf{Predicate}                                                                                                                                                              & \textbf{True branch} & \textbf{Branch-unique computation} \\
		\midrule

		(1) Is the trial rejected?                                                                                                                                                      &
		\textbf{Backtracking line search} shortens the step and retests. This occurs repeatedly until a convergence condition is met.                                                   &
		Re-evaluation of $f$ and $c$, filter acceptance test, reduction of $\alpha$.                                                                                                                                                                \\

		(2) Does the trial have surprising characteristics?                                                                                                                             &
		A \textbf{second-order correction} changes the step direction based on information gathered at the original trial. This occurs repeatedly until a convergence condition is met. &
		Modified right hand side of Equation \eqref{eqn:augmented_kkt}, solved against the \emph{existing} factorization of the left hand side of Equation \eqref{eqn:augmented_kkt}, re-evaluation of $f$ and $c$.                                 \\

		(3) Is a short run of rejected iterates hiding longer-horizon progress?                                                                                                         &
		The \textbf{watchdog} accepts a small sequence of trial steps unconditionally, either accepting the resultant trial, or reverting to the entry iterate.                         &
		Copy and restore of the reference iterate and direction, counters. Scalar and vector arithmetic only.                                                                                                                                       \\

		(4) Are the steps really small?                                                                                                                                                 &
		A \textbf{tiny-step} rule accepts it unconditionally.                                                                                                                           &
		A convergence check using scalar arithmetic at the current trial point only.                                                                                                                                                                \\

		(5) Are there repeated trial rejections?                                                                                                                                        &
		\textbf{Soft feasibility restoration} changes the acceptance criteria, and continues until original criteria are fulfilled or an iteration limit is reached.                    &
		Re-evaluation of $f$, $c$ and their derivatives, acceptance test.                                                                                                                                                                           \\

		(6) Is the optimizer unable to make progress?                                                                                                                                   &
		\textbf{Hard feasibility restoration} solves a new problem whose objective is the constraint violation, returning to the original problem once an acceptable point is found.    &
		None beyond the shared work; the next iteration instantiates the restoration.                                                                                                                                                               \\

		(7) Is the optimization unstable?                                                                                                                                               &
		\textbf{Barrier parameter update} switches from an \textbf{adaptive} rule to a \textbf{monotone} schedule.                                                                      &
		Both updates are computed for every member regardless of the update rule. Scalar arithmetic only.                                                                                                                                              \\

		\bottomrule
	\end{tabularx}
	\caption{\small A simplified overview of \texttt{jaxipm}'s control flow. Branch-unique quantities and predicates are computed unconditionally, but only selected when the predicate is true.}
	\label{tab:branches}
	\vspace{-7mm}
\end{table*}

\noindent In this section, we define the notation used in this paper, the class of optimization problems that we target, and the batched generalization that our optimizer \texttt{jaxipm} is designed to solve. We first introduce a parametric nonlinear program, and then describe a batch as a collection of such programs that share the same structure but differ in their internal parameters.

\noindent \textbf{Notation.} The $i$-th row of a matrix $M \in \mathbb{R}^{n\times m}$ is written as $M_i$. The $i$-th element of a vector $v \in \mathbb{R}^n$ is written as $v^{(i)}$. We define the $\text{diag}$ operator on a vector $v \in \mathbb{R}^n$ as the diagonal matrix $\text{diag}(v) \in \mathbb{R}^{n \times n}$ with $v$ on its diagonal, and zeros elsewhere. We also define $e$ as the vector of all ones for the appropriate dimension, as well as $n_x$ and $n_c$ to be the number of primal optimization variables and equality constraints, respectively.

\noindent \textbf{Single Instance.} We consider parametric nonlinear programs of the form:

\begin{subequations}
	\label{eqn:nlp_full}
	\begin{align}
		\min_{x\in \mathbb{R}^{n_x}} \quad & f(x, \theta) \label{eqn:nlp_full_obj}            \\
		\text{s.t.} \quad                  & c(x, \theta) = 0 \label{eqn:nlp_full_eq}         \\
		                                   & x_L \leq x \leq x_U. \label{eqn:nlp_full_bounds}
	\end{align}
\end{subequations}

\noindent where $x \in \mathbb{R}^{n_x}$ is the decision variable, and $\theta \in \mathbb{R}^{n_\theta}$ is a parameter vector that defines the optimization problem, where $n_\theta$ is the number of parameters. Here, $f \colon \mathbb{R}^{n_x} \times \mathbb{R}^{n_\theta} \to \mathbb{R}$ is the objective, $c \colon \mathbb{R}^{n_x} \times \mathbb{R}^{n_\theta} \to \mathbb{R}^{n_c}$ collects the equality constraints, and $x_L, x_U \in \mathbb{R}^{n_x}$ are elementwise lower and upper bounds on $x$. We assume $f$ and $c$ are twice continuously differentiable in $x$ for every $\theta$ of interest. Note that general inequality constraints can also be included in this formulation through slack variables.

\noindent \textbf{Batched Instances.} A batch collects $N$ instances of problem~\eqref{eqn:nlp_full} that share the functions $f$ and $c$ and the bounds $x_L$, $x_U$, and that differ only in the parameter $\theta$. We write the decision variable and parameter of the $i$-th instance as $X_i \in \mathbb{R}^{n_x}$ and $\Theta_i \in \mathbb{R}^{n_\theta}$, respectively. Here, $X \in \mathbb{R}^{N \times n_x}$ is the stack of decision variables for the batch, and $\Theta \in \mathbb{R}^{N \times n_\theta}$ is the stack of parameters for the batch. We can now define the batched problem:

\begin{equation}
	\left.
	\begin{aligned}
		\min_{X_i \in \mathbb{R}^{n_x}} & \; f(X_i, \Theta_i)               \\
		\text{s.t.}                     & \; c(X_i, \Theta_i) = 0           \\
		                                & \; X_{i_L} \leq X_i \leq X_{i_U},
	\end{aligned}
	\right\} \quad \forall i \in \{1, \dots, N\}
	\label{eqn:nlp_batch_naive}
\end{equation}

\noindent where each row $X_i$ is solved as an independent program. Here, $X_{i_L}$ and $X_{i_U}$ are the lower and upper bounds for the $i$-th instance, which equal the shared bounds $x_L$ and $x_U$ for every $i$. Instances are decoupled in their decision variables, so a batch of $N$ programs is mathematically equivalent to $N$ separate calls to a single-instance optimizer.
Our approach, \texttt{jaxipm}, is built to exploit this structure on GPU hardware\footnote{For more information on programming for GPU hardware we refer the reader to \cite{gspmd}.} by adapting IPOPT, one of the most widely used solver for problems of the form \eqref{eqn:nlp_full}. We review the IPOPT algorithm in the next section before detailing our contributions.

\section{IPOPT Background}

\noindent In this section, we provide a brief background to IPOPT for solving~\eqref{eqn:nlp_full} and explain {its difficulties in scaling to} solve~\eqref{eqn:nlp_batch_naive}.
We first introduce how it calculates step directions, then its techniques for handling edge cases throughout its optimization, and finally explain why this makes {a single IPOPT instance} ineffective at solving the batched problem~\eqref{eqn:nlp_batch_naive}.
When solving \eqref{eqn:nlp_full}, for simplicity, we assume that every $x^{(i)}$ is upper and lower bounded by $x_L^{(i)}$, $x_U^{(i)}$\footnote{Note that the changes necessary to handle the one-sided bounds are outlined in Section 3.4 in \cite{ipopt}.}. Since the parameters $\theta$ are fixed within a single solve, we suppress them and write $f(x)$ and $c(x)$ for $f(x, \theta)$ and $c(x, \theta)$.

\noindent \textbf{Prerequisites.} IPOPT is a primal-dual interior-point method. Like \cite{trust_region_ipm, trust_region_pd, nonlinear_programming}, it enforces the bound constraints \eqref{eqn:nlp_full_bounds} with a logarithmic barrier, forming the barrier objective:
\begin{equation*}
	\varphi_\mu(x) := f(x) - \mu \sum_{i=1}^{n_x} \ln (x^{(i)} - x_L^{(i)}) - \mu \sum_{i=1}^{n_x}\ln(x_U^{(i)} - x^{(i)}),
\end{equation*}

\noindent with barrier parameter $\mu > 0$, and solves a sequence of barrier problems while driving $\mu$ to zero \cite{ipopt}.
It maintains a primal-dual iterate $(x, y_c, z_L, z_U)$ where $y_c$, $z_L$ and $z_U$ are the Lagrange multipliers for \eqref{eqn:nlp_full_eq}, the lower bounds, and the upper bounds of \eqref{eqn:nlp_full_bounds}, respectively.
The Lagrangian is
\begin{equation*}
	\mathcal{L}(x, y_c, z_L, z_U) := f(x) + y_c^\top c(x) - z_L^\top (x - x_L) - z_U^\top (x_U - x).
\end{equation*}
\noindent To compute a step, IPOPT applies Newton's method to its optimality conditions:
\begin{equation}
	\begin{bmatrix}
		W + \Sigma & A^\top \\
		A          & 0
	\end{bmatrix}
	\begin{bmatrix} \Delta x \\ \Delta y_c \end{bmatrix}
	= -\begin{bmatrix} \nabla \varphi_\mu(x) + A^\top y_c \\ c(x) \end{bmatrix}.
	\label{eqn:augmented_kkt}
\end{equation}

\noindent where $W := \nabla_{xx}^2 \mathcal{L}(x, y_c, z_L, z_U)$ denotes the Lagrangian Hessian and $A := \nabla c(x)$ the constraint Jacobian.\footnote{The remaining steps $\Delta z_L, \Delta z_U$ can be recovered from equations detailed in Section 2.2 in \cite{ipopt}.}

\noindent \textbf{Edge Case Handling.} The Newton step from \eqref{eqn:augmented_kkt} may be a poor choice of direction, leading to infeasibility or away from a local minimum.
This is why IPOPT and \texttt{jaxipm} have several fallback mechanisms for common failure modes, which we list in Table \ref{tab:branches}.
For example, row (1) of Table \ref{tab:branches} asks: \textit{Is the trial} (Newton step) \textit{rejected}? This typically means that the landscape of the optimality conditions is rapidly changing, and so the linearization used to form \eqref{eqn:augmented_kkt} is only locally accurate.
To combat this, the size of the step is reduced according to a backtracking scheme, bringing the step closer to where the linearization was accurate. Similarly, there are other failure modes with designed fallbacks that we provide some intuition for in Table \ref{tab:branches}.

\begin{figure}[t]
	\centering
	\includegraphics[width=\linewidth]{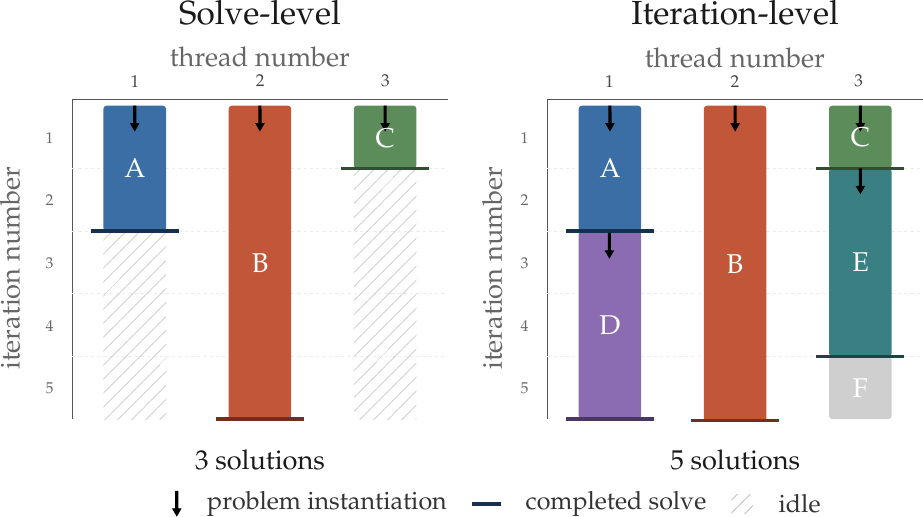}
	\caption{\small We visualize the throughput advantages of iteration-level batching versus solve-level batching. Left: We show solve-level batching achieving 3 solutions (optimizations A, B, and C) within 5 iterations, with substantial compute idle time. Right: We show iteration-level batching achieving 5 solutions (optimizations A, B, C, D, and E) within 5 iterations, with no compute idle time and a one-iteration head start on optimization F.}
	\label{fig:iter_lvl_batch}
\end{figure}

\noindent\textbf{Why Is IPOPT Hard To Batch?} What we have spoken about so far is how IPOPT goes about solving a single problem.
If we want to solve multiple problems concurrently, as is the case in \eqref{eqn:nlp_batch_naive}, a naive approach would stack multiple problems into a \emph{joint} batch problem and use a single instance of IPOPT to solve.
This approach, however, prevents IPOPT from selecting different control-flow branches (Table \ref{tab:branches}) for different problems as it is oblivious to the internal (decoupled) structure of the batch.
At best, this would inflate iteration counts; at worst, it would cause individual solves to fail.
An alternative is utilizing \emph{concurrent} IPOPT instances across multiple CPU cores, which retains independent branching but ties throughput to core count and keeps solves off the GPU.
In this work, we instead seek to retain IPOPT’s per-problem branching within a single GPU-batched solver.

\section{Batched Interior-Point Method}

\noindent In this section, we will describe how we rebuilt IPOPT from scratch to enable high throughput batch solving whilst maintaining per-optimization control flow.
We will go over how we fused all control-flow branches via \textit{heterogeneous iteration fusion}, and then how we accelerated throughput via \textit{iteration-level batching}.

\noindent \textbf{Heterogeneous Iteration Fusion.}
IPOPT does not effectively solve \eqref{eqn:nlp_batch_naive} directly because it cannot select different control-flow branches for $X_i$ and $X_j$ when $i \neq j$.
Consequently, even when $X_i$ and $X_j$ would benefit from different control-flow branches, which we call heterogeneous iterations, they are forced to take the same branch.
This problem would be solved if we could somehow remove all control-flow branches so that every optimization in the batch always did the same operation.
We have done exactly this in \texttt{jaxipm} by walking through each control-flow branch in \cref{tab:branches} and splitting its work into two parts: computation that is shared across branches and computation that is unique to branches. 
Shared work is computed once per iteration.
For the remaining branch-unique work, we compute all branches for all optimizations and select from the outputs based on which branch was taken.
Quantities unconditionally evaluated by IPOPT include the problem functions and their derivatives, the Newton step, and the fraction-to-boundary rule.
Therefore, these computations are trivially classed as shared work we can compute concurrently across the batch.

Certain operations take further work to homogenize, such as the hard feasibility restoration.
In IPOPT, this branch will instantiate a nested IPOPT optimization problem which aims to minimize constraint violation as opposed to the regular objective.
This second optimization problem can itself contain all branches in \cref{tab:branches}.
To avoid these branches in \texttt{jaxipm}, we homogenized the computation graph of the regular and restoration optimization problems through careful padding of their decision variables. 
This enables concurrent computation of both iteration types.
Similarly, the second-order correction and backtracking line search in \cref{tab:branches} need further work as both contain loops which terminate conditionally.
To homogenize these branches, \texttt{jaxipm} iterates these loops across all members of the batch until \textit{all} members reach their termination condition.

\noindent \textbf{Iteration-Level Batching.} What we have discussed so far still leaves one large inefficiency: the batch only advances as fast as its slowest member.
For example, if one problem converges in 10 iterations and another in 100, then the faster problem must wait for the slower one to complete before the next batch can begin.
Rather than letting the throughput of our batch optimizer be determined by the pace of its slowest member, upon convergence of one member, we start a new optimization in its place.
When a member converges, we initialize a fresh problem in its place on the next iteration, while all other members continue iterating uninterrupted. See \cref{fig:iter_lvl_batch} for a graphical illustration of this.
The fresh problem initialization is done in exactly the same way as IPOPT would instantiate a new problem from a new parameterization and initial condition.
For further details on problem initialization, we refer the reader to~\cite[Section 3.6]{ipopt}. 
This means that we \textit{batch} at the \textit{iteration level} rather than the solve level.

\section{Results}

This section presents our results.
We create three NMPC scenarios designed to demonstrate our optimizer \texttt{jaxipm} on robotics scenarios.
We first demonstrate raw \textit{throughput}, and the benefit of iteration-level batching.
After testing nominial throughput, we also test \texttt{jaxipm}'s sensitivity to problem difficulty and dimensionality.
Finally, we evaluate \textit{correctness} against IPOPT. All tests are performed using a single NVIDIA L40S (48 GB VRAM) and a 64-core dual-socket AMD EPYC 9354 (378 GB RAM).
We run all tests at FP64 precision using a convergence tolerance of $10^{-8}$ and an iteration limit of~$500$.

\subsection{Experimental Setup}

\noindent\textbf{Shared Problem Setup.}
We use quadrotors in all of these experiments due to their nonlinear, second-order-differentiable dynamics.
We consider 13 DoF quadrotors whose state~$x = (p, q, v, \omega) \in \mathbb{R}^3 \times \mathbb{R}^4 \times \mathbb{R}^3 \times \mathbb{R}^3 $ stacks position, attitude quaternion, and body linear and angular velocities.
The dynamics $\dot{x} = f(x,u)$ combine standard quadrotor translational and rotational dynamics~\cite{quadrotor_dynamics} with four rotor angular velocity inputs $u \in \mathbb{R}^4$.
In all tests, we use NMPC to perform trajectory optimization on quadrotors with a timestep of~0.1s and a prediction horizon of~30.
This means that each quadrotor optimization has a minimum of~506 optimization variables, excluding slacks from inequality constraints.
Although the quadrotor represents a small fraction of robotics problems our general-purpose NLP optimizer can handle, it demonstrates the benefits of our contributions in this paper: iteration-level batching and heterogeneous iteration fusion.
We also demonstrate algorithmic parity with IPOPT, and therefore claim that our method can be used wherever IPOPT has been used before.
With this in mind, we have designed a set of experiments centered around quadrotors that stress test \textit{throughput} under various levels of problem difficulty and dimensionality, and \textit{correctness}.

\begin{table}[b]
	\centering
	\setlength{\tabcolsep}{2.5pt}
	{\scriptsize
		\begin{tabular}{@{}l r r r r r@{}}
			\toprule
			Solver               & solves/s           & \makecell{equiv.                                           \\instances} & \makecell{success\\ (\%)} & cost & \makecell{iters\\ mean/max} \\
			\midrule
			IPOPT-default        & 6.99\,$\pm$\,0.28  & 1.00             & 100   & 507.6\,$\pm$\,17.1 & 21.5 / 71  \\
			IPOPT-64             & 310.9\,$\pm$\,14.8 & 44.5             & 100   & 507.6\,$\pm$\,17.1 & 21.5 / 71  \\
			MadNLP               & 2.74\,$\pm$\,0.02  & 0.39             & 100   & 507.6\,$\pm$\,17.2 & 32.2 / 184 \\
			\jaxipmnoIL & 14.08\,$\pm$\,1.12 & 2.01             & 99.55 & 507.6\,$\pm$\,17.2 & 71.6 / 468 \\
			\texttt{jaxipm}      & 95.45\,$\pm$\,1.24 & 13.7             & 100   & 507.6\,$\pm$\,16.5 & 51.9 / 252 \\
			\bottomrule
		\end{tabular}}
	\caption{\small We demonstrate our solver's throughput advantage over {IPOPT-default and MadNLP} in terms of solves per second. We also note the average final cost function value and iteration count of every optimization to demonstrate that the quality of our solutions is comparable with baselines in batch, and we do not require excessive iterations to achieve them.}
	\label{tab:nav-throughput}
	\vspace{-1em}
\end{table}

\noindent \textbf{Compared Methods.}
We compare against CasADi/IPOPT \cite{casadi, ipopt} and ExaModels/MadNLP \cite{madnlp}.
In all benchmarks we run IPOPT in two ways.
\emph{IPOPT-default} solves optimization problems one by one{, with its linear solver (MUMPS) multithreaded across} all 64 cores of our CPU.
This baseline serves as a reference for single-instance throughput under ideal conditions.
\emph{IPOPT-64} concurrently solves 64 problem instances{, each pinned to one physical CPU core}.
This baseline serves as a reference for batch parallelism when using the full capacity of a high-end CPU.
MadNLP is a GPU-\textit{accelerated} version of IPOPT that utilizes cuDSS as its linear solver.
Since MadNLP requires a GPU for each problem instance, we only include it as a sequential variant in all benchmarks.
For our approach, we evaluate two variants: \emph{\texttt{jaxipm}} is our default method and \emph{\jaxipmnoIL} removes iteration-level batching.
For both of these variants we select batch sizes to fully saturate the GPU.
In all benchmarks, each optimization problem within a batch has its own parameterization and initial condition provided to the optimizer.
These parameterization and initial condition pools are common between all optimizers to keep all experiments fair.
The number of problem instances in each benchmark is $4\times$ \texttt{jaxipm}'s batch size.

\noindent \textbf{Metrics.}
We report ``solves/s'' as a metric of throughput, ``success'' percentage as a metric for robustness, and ``cost'' for solution quality.
For additional insights we report iteration count statistics (``iters mean/max''), and the speedup relative to IPOPT-default (``equiv. instances'').

\begin{figure}[t]
	\centering
	\includegraphics[width=\linewidth]{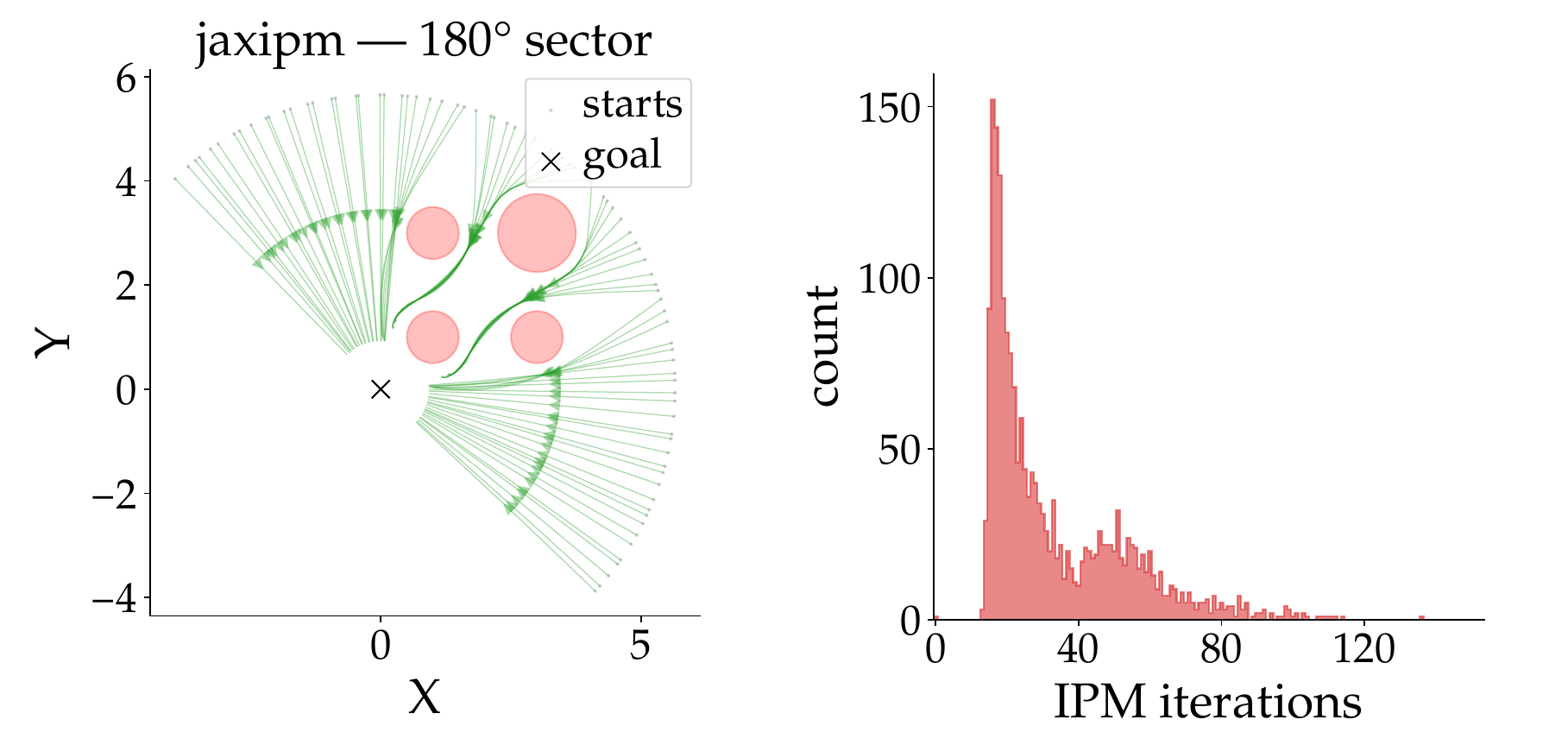}
	\vspace{-2em}
	\caption{\small \textbf{Left:} The top-down view of a batch of optimization solutions (80/2000) from different initial positions, all of which must navigate to a common goal position whilst avoiding obstacles. We initialize quadrotors in a $180^\circ$ arc around the destination. \textbf{Right:}~The corresponding iteration count distribution.}
	\label{fig:throughput_iter_lvl}
\end{figure}

\subsection{What throughput does \texttt{jaxipm} achieve?}
\label{sec:result-throughput}

\noindent\textbf{Experimental Setup.}
We first test throughput by setting up a batch of quadrotors that must each navigate to a common position from varying initial positions.
The initial positions are drawn from a circular arc of $180^\circ$ as can be seen in \cref{fig:throughput_iter_lvl}.
This choice of initial conditions combined with the obstacle layout leads to a batch of problems with a high variance of difficulty: some batch instances are solved by a trivial, near-straight trajectory to the goal, while others must navigate around multiple obstacles to reach their goal.
Therefore, this experiment design tests the optimizers' throughput when facing batches of problems with heterogeneous difficulty.
The multi-modal distribution of iteration counts per problem instance shown in \cref{fig:throughput_iter_lvl} highlights this diversity in problem hardness.
In this test \texttt{jaxipm} and \jaxipmnoIL use a batch size of $500$.

\noindent\textbf{Results: Baseline Comparison.}
\Cref{tab:nav-throughput} shows the results for this experiment.
We observe that \texttt{jaxipm}'s throughput is equivalent to that of $13.7$~IPOPT-default instances, making it $34.8\times$~faster than the GPU-accelerated MadNLP baseline.
In fact, \texttt{jaxipm} (running on an L40S GPU) achieves $30.7\%$ of the throughput of IPOPT-64 (running on all 64 CPU cores of a dual-socket AMD EPYC 9354).

\noindent\textbf{Results: Ablation Comparison.}
Furthermore, observe that our method's performance strongly relies on the iteration-level batching feature.
Without it (\jaxipmnoIL), the throughput drops to only~$14.8\%$ of our method with all features enabled.
This ablation highlights the utility of contribution~\textbf{(C2)}.

\subsection{How does problem difficulty affect \texttt{jaxipm}?}
\label{sec:result-difficulty}
\begin{figure}[t]
	\centering
	\includegraphics[width=0.7\linewidth]{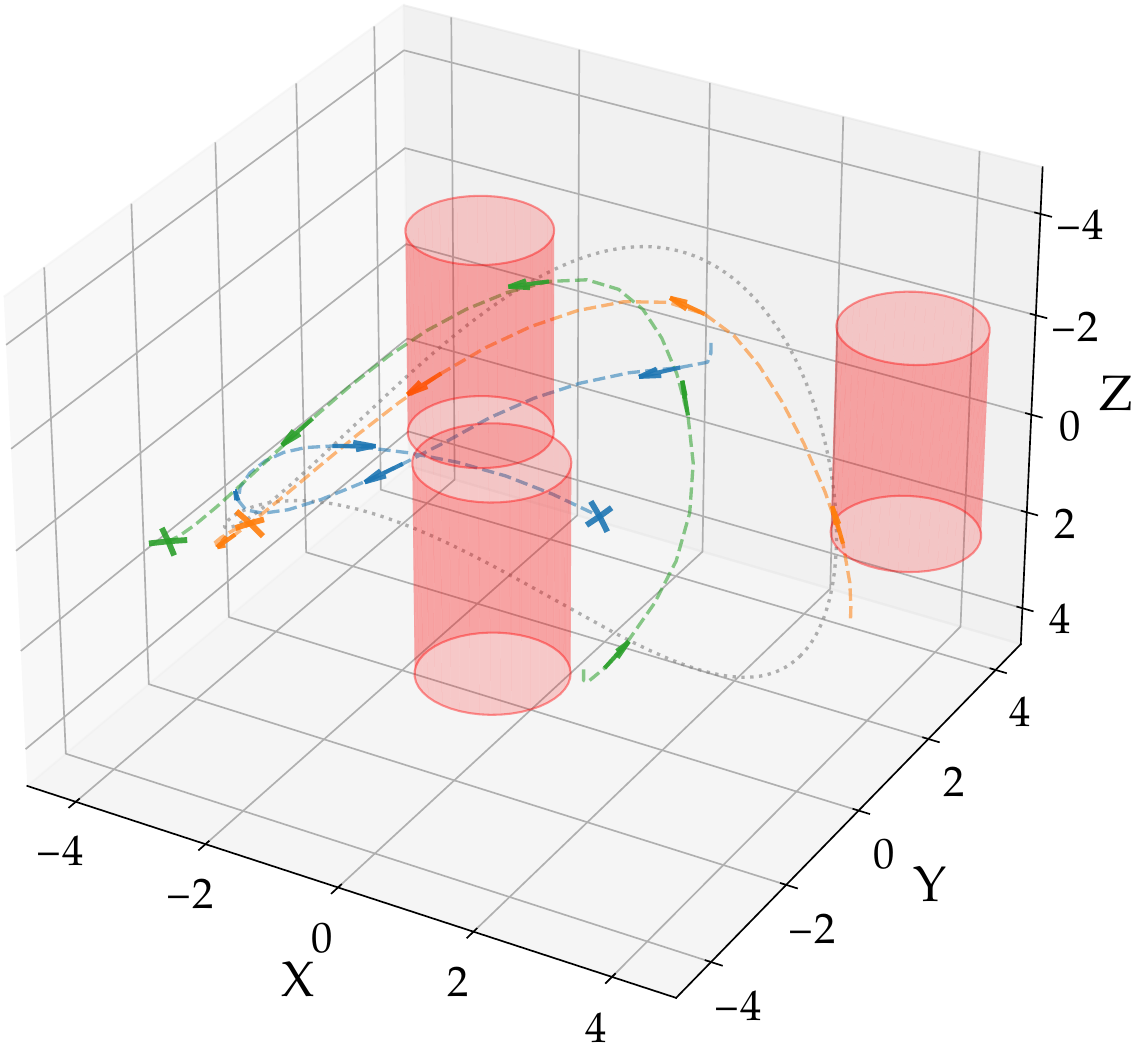}
	\caption{\small We visualize the reference tracking with time-varying obstacles task at $\bar{v} = 1.5$ m/s. %
	}
	\label{fig:track_avoid}
\end{figure}

\noindent\textbf{Experimental Setup.}
Next, we test how problem difficulty affects each optimizer's throughput.
To this end, we study a quadrotor reference-tracking task with time-varying obstacles whose trajectories are known a priori.
In this setting, we generate problem variants of varying difficulty by increasing the reference velocity being tracked $\bar{v}$.
\Cref{fig:track_avoid} visualizes the experiment setup.
Each batch instance tracks the same periodic reference but starts at a different phase along the reference.
In this test \texttt{jaxipm} and \jaxipmnoIL use a batch size of $250$.

\noindent\textbf{Results: Baseline Comparison.}
\Cref{tab:track-avoid-throughput} shows the results of this experiment.
We observe that \texttt{jaxipm}'s throughput is equivalent to that of $15.3$~to~$28.9$~IPOPT-default instances, leaving our approach $31.1\times$~to~$44.7\times$ faster than the GPU-accelerated MadNLP baseline.
Compared to the IPOPT-64 baseline that uses the full~CPU, \texttt{jaxipm} achieves $39.4\%$~to~$69.9\%$ of its throughput on a single GPU.
Another observation we make is that \texttt{jaxipm}'s iteration count remains relatively constant across difficulties.
We attribute this effect to the batch linear solver used by \texttt{jaxipm}.
Specifically, the batch linear solver we use~\cite{cudss} cannot reliably apply iterative refinement when operating in batch.
Empirically, we find that these more inaccurate linear system solutions impact the solution time of ``easy'' problems more than that of ``hard'' ones.
Finally, we remark that \texttt{jaxipm} achieves a different average cost from the other optimizers.
We verified that these are simply different local minima that happen to be found by the inaccurate directions used by~\texttt{jaxipm}.

\noindent\textbf{Results: Ablation Comparison.}
We also observe that the success rate of \jaxipmnoIL diminishes to $86.1\%$ when facing harder problem instances, while \texttt{jaxipm} remains at $100\%$~success rate.
This is because when \jaxipmnoIL freezes solved systems, it still wastefully computes iterations for them as the rest of the batch converges.
The final converged iteration for interior-point methods is well known to be very poorly conditioned \cite{poor_conditioned_ipm}.
A limitation of the batch linear solver used by \texttt{jaxipm}~\cite{cudss} is that poor conditioning in one batch element can corrupt others, amplifying the inaccuracy of the step directions calculated for the long-running optimizations when performing solve-level batching.
We note that \texttt{jaxipm} (with iteration-level batching enabled) does not experience this because replacing completed problems has the side effect of removing these ill-conditioned linear systems from the batch.
This highlights another benefit of contribution~\textbf{(C2)}: iteration-level batching not only improves throughput, but also robustness.

\begin{table}[t]
	\centering
	\setlength{\tabcolsep}{2.5pt}
	{\scriptsize
		\begin{tabular}{@{}l r r r r r@{}}
			\toprule
			Solver               & solves/s           & \makecell{equiv.                                            \\instances} & \makecell{success\\ (\%)} & cost & \makecell{iters\\ mean/max} \\
			\midrule
			\multicolumn{6}{@{}l}{\textit{$\bar{v} = 1.0$ m/s}}                                                     \\
			IPOPT-default        & 6.06\,$\pm$\,0.30  & 1.00             & 100  & 265.7\,$\pm$\,98.7   & 25.4 / 77  \\
			IPOPT-64             & 234.7\,$\pm$\,9.5  & 38.7             & 100  & 265.7\,$\pm$\,98.7   & 25.4 / 77  \\
			MadNLP               & 2.97\,$\pm$\,0.99  & 0.49             & 100  & 261.7\,$\pm$\,92.0   & 38.0 / 84  \\
			\jaxipmnoIL & 13.00\,$\pm$\,0.71 & 2.15             & 99.2 & 249.8\,$\pm$\,79.4   & 54.2 / 472 \\
			\texttt{jaxipm}      & 92.49\,$\pm$\,0.37 & 15.3             & 100  & 244.0\,$\pm$\,76.3   & 43.1 / 181 \\
			\midrule
			\multicolumn{6}{@{}l}{\textit{$\bar{v} = 1.5$ m/s}}                                                     \\
			IPOPT-default        & 4.61\,$\pm$\,0.27  & 1.00             & 100  & 556.7\,$\pm$\,140.9  & 30.3 / 71  \\
			IPOPT-64             & 185.4\,$\pm$\,3.8  & 40.2             & 100  & 556.7\,$\pm$\,140.9  & 30.3 / 71  \\
			MadNLP               & 2.02\,$\pm$\,0.01  & 0.44             & 100  & 555.8\,$\pm$\,138.3  & 47.4 / 116 \\
			\jaxipmnoIL & 11.85\,$\pm$\,0.07 & 2.57             & 96.2 & 543.2\,$\pm$\,126.5  & 54.3 / 494 \\
			\texttt{jaxipm}      & 81.76\,$\pm$\,0.37 & 17.7             & 100  & 539.2\,$\pm$\,121.4  & 44.4 / 237 \\
			\midrule
			\multicolumn{6}{@{}l}{\textit{$\bar{v} = 2.0$ m/s}}                                                     \\
			IPOPT-default        & 2.55\,$\pm$\,0.14  & 1.00             & 100  & 1146.7\,$\pm$\,126.9 & 47.8 / 125 \\
			IPOPT-64             & 105.4\,$\pm$\,3.7  & 41.3             & 100  & 1146.7\,$\pm$\,126.9 & 47.8 / 125 \\
			MadNLP               & 1.65\,$\pm$\,0.00  & 0.65             & 100  & 1147.5\,$\pm$\,121.2 & 57.8 / 136 \\
			\jaxipmnoIL & 11.32\,$\pm$\,0.22 & 4.44             & 86.1 & 1140.5\,$\pm$\,125.3 & 46.9 / 165 \\
			\texttt{jaxipm}      & 73.71\,$\pm$\,0.87 & 28.9             & 100  & 1142.4\,$\pm$\,131.5 & 45.8 / 129 \\
			\bottomrule
		\end{tabular}}
	\caption{\small Per-solve throughput and solution quality on the quadrotor reference-tracking problem with three time-varying obstacles. We control difficulty by varying the average velocity $\bar{v}$ of the reference trajectory.}
	\label{tab:track-avoid-throughput}
	\vspace{-2em}
\end{table}

\subsection{How does dimensionality affect \texttt{jaxipm}?}
\label{sec:result-dimensionality}
\begin{figure}[t]
	\centering
	\includegraphics[width=\linewidth]{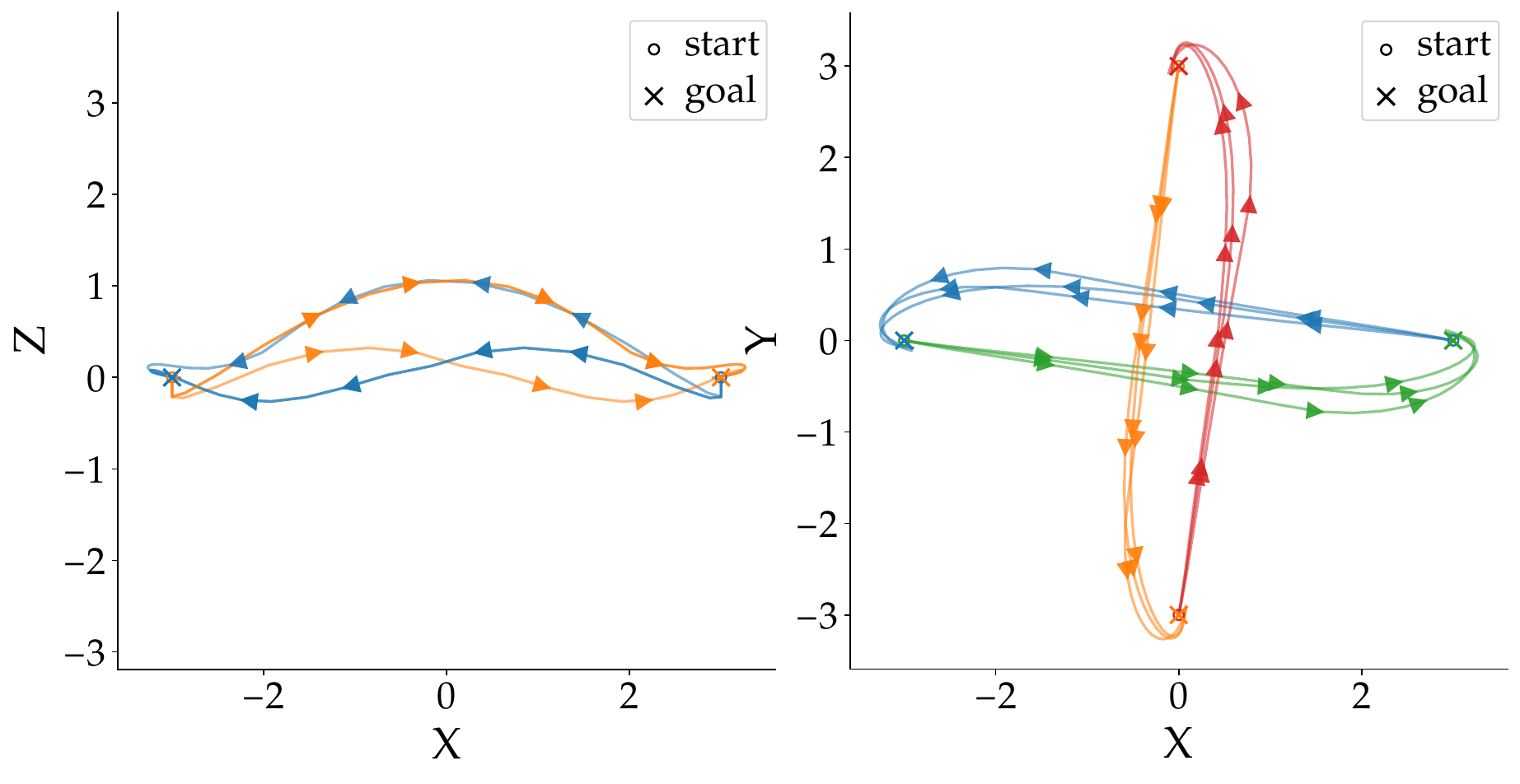}
	\caption{\small We visualize the multi-quadrotor consensus task running on our method with 2 and 4 quadrotors. Each colored trajectory shows one quadrotor's path from its start to its assigned goal. The quadrotors must swap positions while maintaining a minimum separation distance with every other quadrotor throughout.}
	\label{fig:multi_quad}
	\vspace{-2em}
\end{figure}

\noindent\textbf{Experimental Setup.}
Next, we test how problem dimensionality affects each optimizer's throughput.
To this end, we study a multi-quadrotor consensus task, where multiple quadrotors are tasked to switch positions whilst maintaining a minimum separation distance with every other quadrotor throughout.
The initial and final conditions for every quadrotor in this scenario are uniformly sampled along a $10^\circ$ arc.
To generate problem variants of different dimensionality, we instantiate the problem with~2 and~4 quadrotors as seen in \cref{fig:multi_quad}.
The problem variants result in~1012 and~2024~optimization variables, respectively. %
In this test \texttt{jaxipm} and \jaxipmnoIL use a batch size of~$75$.

\noindent\textbf{Results: Baseline Comparison.}
\Cref{tab:multi-swap-throughput} shows the results of this experiment.
Here, \texttt{jaxipm}'s throughput is equivalent to that of $15.7$~to~$31.9$~IPOPT-default instances, making our approach $7.62\times$~to~$17.2\times$ faster than the GPU-accelerated MadNLP baseline.
Compared to the IPOPT-64 baseline, \texttt{jaxipm} achieves $32.5\%$~to~$69.8\%$ of its throughput.
We find that \texttt{jaxipm}'s throughput decreases relative to the baseline as the problem dimensionality increases.
This is because fewer \emph{large} problems fit into the limited GPU memory, and so the GPU saturates at a smaller batch size.

\noindent\textbf{Results: Ablation Comparison.}
Compared to the experiments in \cref{sec:result-throughput,sec:result-difficulty}, we observe that the throughput advantage of iteration-level batching is less pronounced in this experiment.
We attribute this to the fact that the problems in this experiment are more homogeneous in difficulty due to the narrow distribution of initial conditions.
As a result, the batch is less likely to be slowed down by a single slow member and the benefit of iteration-level batching is less pronounced.

\begin{table}[b]
	\centering
	\setlength{\tabcolsep}{2.5pt}
	{\scriptsize
		\begin{tabular}{@{}l r r r r r@{}}
			\toprule
			Solver               & solves/s            & \makecell{equiv.                                          \\instances} & \makecell{success\\ (\%)} & cost & \makecell{iters\\ mean/max} \\
			\midrule
			\multicolumn{6}{@{}l}{\textit{2 quadrotors}}                                                           \\
			IPOPT-default        & 0.578\,$\pm$\,0.018 & 1.00             & 100 & 755.5\,$\pm$\,0.8  & 112.6 / 135 \\
			IPOPT-64             & 26.38\,$\pm$\,0.98  & 45.6             & 100 & 755.5\,$\pm$\,0.8  & 112.7 / 135 \\
			MadNLP               & 1.07\,$\pm$\,0.14   & 1.84             & 100 & 755.5\,$\pm$\,0.8  & 127.7 / 162 \\
			\jaxipmnoIL & 17.31\,$\pm$\,0.46  & 29.9             & 100 & 755.5\,$\pm$\,0.8  & 115.7 / 176 \\
			\texttt{jaxipm}      & 18.42\,$\pm$\,0.20  & 31.9             & 100 & 755.5\,$\pm$\,0.8  & 113.6 / 182 \\
			\midrule
			\multicolumn{6}{@{}l}{\textit{4 quadrotors}}                                                           \\
			IPOPT-default        & 0.289\,$\pm$\,0.008 & 1.00             & 100 & 1529.7\,$\pm$\,4.7 & 95.5 / 159  \\
			IPOPT-64             & 13.99\,$\pm$\,0.13  & 48.4             & 100 & 1529.7\,$\pm$\,4.7 & 93.3 / 140  \\
			MadNLP               & 0.596\,$\pm$\,0.008 & 2.06             & 100 & 1529.7\,$\pm$\,4.7 & 184.1 / 217 \\
			\jaxipmnoIL & 2.62\,$\pm$\,0.03   & 9.06             & 95  & 1533.9\,$\pm$\,8.3 & 241.7 / 498 \\
			\texttt{jaxipm}      & 4.54\,$\pm$\,0.05   & 15.7             & 100 & 1533.7\,$\pm$\,8.9 & 227.3 / 484 \\
			\bottomrule
		\end{tabular}}
	\caption{\small Per-solve throughput and solution quality on the multi-quadrotor consensus problem, where every quadrotor maintains a minimum separation from every other in the fleet.}
	\label{tab:multi-swap-throughput}
\end{table}

\subsection{Does \texttt{jaxipm} produce the same results as IPOPT?}
\label{sec:result-correctness}

\noindent\textbf{Experimental Setup.}
Our implementation of \texttt{jaxipm} is thoroughly unit-tested to ensure a faithful recreation of IPOPT's algorithmic behavior on the GPU.
Beyond these unit tests, however, here we present one additional test to validate our algorithmic parity with IPOPT.
Specifically, we run IPOPT to convergence from a single challenging starting point ($x=4$, $y=4$) in the navigation task of~\cref{fig:throughput_iter_lvl}.
For this test, we configure cuDSS with conservative iterative-refinement settings, allowing many refinement iterations, to compare \texttt{jaxipm} and IPOPT at comparable linear-solve accuracy.\footnote{This conservative cuDSS setting relies on the fact that, here, we solve a \emph{single} problem instance. Iterative refinement is not reliable in batch, as discussed in \cref{sec:result-difficulty}.}
Among all tests in this section, this optimization goes through the most control-flow branches of IPOPT.
These include second-order correction, soft/hard feasibility restoration, and free/monotone barrier update switches (see Table \ref{tab:branches}).
We save the complete IPOPT state at every iteration.
We then load this complete state into \texttt{jaxipm} and step \texttt{jaxipm} once.
We then convert the new \texttt{jaxipm} state and the next IPOPT state into vectors $v^\texttt{jaxipm}$ and $v^\text{IPOPT}$, respectively.
We then find the largest scaled elementwise deviation $e_v$ using the following formula (restricted to quantities IPOPT computes; fields with no IPOPT counterpart are excluded):
\begin{equation}
	e_v \;=\; \max_i \;
	\frac{\bigl\lvert v_i^{\texttt{jaxipm}} - v_i^{\text{IPOPT}} \bigr\rvert}
	{1 + \bigl\lvert v_i^{\text{IPOPT}} \bigr\rvert}.
\end{equation}
By normalizing the deviation by $1 + \bigl\lvert v_i^{\text{IPOPT}} \bigr\rvert$, this metric measures the relative deviation without a singularity at zero.\footnote{This error metric underpins \texttt{numpy.isclose} when $\texttt{rtol}==\texttt{atol}$.}

\begin{figure}[t]
	\centering
	\includegraphics[width=\linewidth]{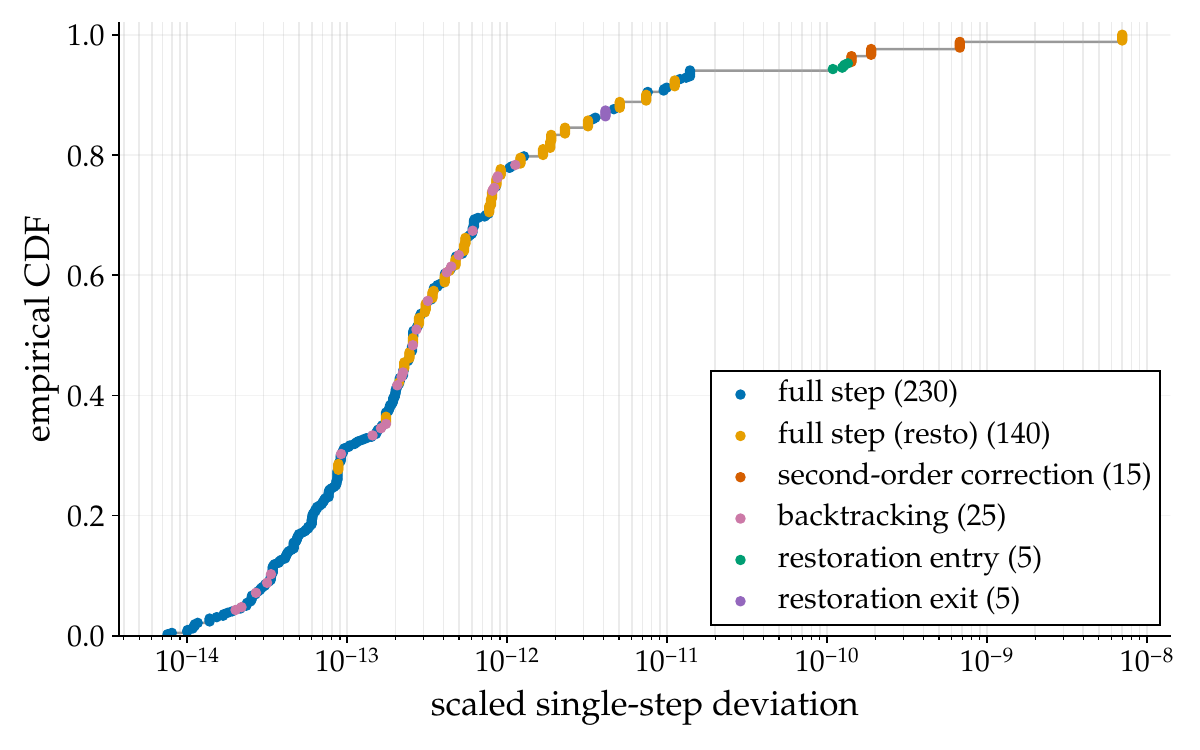}
	\caption{\small We visualize the ECDF of the step-wise discrepancies between our method \texttt{jaxipm} and IPOPT across five full optimizations.}
	\label{fig:ecdf}
	\vspace{-1em}
\end{figure}

\noindent\textbf{Results.}
\Cref{fig:ecdf} shows the empirical cumulative distribution function (ECDF) of the step-wise deviations between \texttt{jaxipm} and IPOPT across five full optimizations.
We observe that all deviations fall beneath $10^{-8}$, which is the convergence tolerance we set for both solvers.
This demonstrates that \texttt{jaxipm} produces results that are numerically equivalent to IPOPT across all control-flow branches activated in this optimization.
In sum, this test provides empirical evidence for \texttt{jaxipm} being a feature-complete recreation of IPOPT, built from the ground up with GPU-batching in mind.

\section{Conclusion}

We present \texttt{jaxipm}, the first GPU-batched NLP optimizer, which can run \emph{hundreds} of independent IPOPT optimizations concurrently on a single GPU.
Our two contributions that enable this are \emph{heterogeneous iteration fusion}, which collapses IPOPT's branching recovery mechanisms into a single iteration type for all optimizations, and \emph{iteration-level batching}, which minimizes GPU idle time.
We evaluate \texttt{jaxipm}'s throughput, scalability, and correctness across a variety of NMPC quadrotor tasks and outperform MadNLP and single-instance IPOPT in terms of throughput on every problem tested.
Compared to IPOPT-64, which concurrently solves problems on every core of a 64-core dual-socket CPU, \texttt{jaxipm} achieves a substantial fraction of its throughput on a single GPU.
We see several future directions for \texttt{jaxipm}.
For example, imitation-learning pipelines could leverage \texttt{jaxipm} to build datasets of locally optimal trajectories directly on the GPU.
Furthermore, GPU-enabled sampling-based techniques such as MPPI or RL could embed constraints per sample via batch-solved NLPs.

\bibliographystyle{IEEEtran}

\balance
\bibliography{lib}

@article{bhatt2025strategic,
  title={Strategic decision-making in multi-agent domains: A weighted constrained potential dynamic game approach},
  author={Bhatt, Maulik and Jia, Yixuan and Mehr, Negar},
  journal={IEEE Transactions on Robotics},
  year={2025},
  publisher={IEEE}
}

@article{traj_opt_nlp,
  title={An introduction to trajectory optimization: How to do your own direct collocation},
  author={Kelly, Matthew},
  journal={SIAM review},
  volume={59},
  number={4},
  pages={849--904},
  year={2017},
  publisher={SIAM}
}

@article{gspmd,
  title={Gspmd: general and scalable parallelization for ml computation graphs},
  author={Xu, Yuanzhong and Lee, HyoukJoong and Chen, Dehao and Hechtman, Blake and Huang, Yanping and Joshi, Rahul and Krikun, Maxim and Lepikhin, Dmitry and Ly, Andy and Maggioni, Marcello and others},
  journal={arXiv preprint arXiv:2105.04663},
  year={2021}
}

@article{poor_conditioned_ipm,
  title={Implementation of interior-point methods for LP based on Krylov subspace iterative solvers with inner-iteration preconditioning: Y. Cui et al.},
  author={Cui, Yiran and Morikuni, Keiichi and Tsuchiya, Takashi and Hayami, Ken},
  journal={Computational optimization and applications},
  volume={74},
  number={1},
  pages={143--176},
  year={2019},
  publisher={Springer}
}

@article{nmpc_nlp,
  title={Nonlinear model predictive control: From theory to application},
  author={Allgower, Frank and Findeisen, Rolf and Nagy, Zoltan K and others},
  journal={Journal-Chinese Institute Of Chemical Engineers},
  volume={35},
  number={3},
  pages={299--316},
  year={2004},
  publisher={CHINESE INST CHEM ENGINEERS}
}

@inproceedings{ik_nlp,
  title={An experimental evaluation of a novel minimum-jerk cartesian controller for humanoid robots},
  author={Pattacini, Ugo and Nori, Francesco and Natale, Lorenzo and Metta, Giorgio and Sandini, Giulio},
  booktitle={2010 IEEE/RSJ international conference on intelligent robots and systems},
  pages={1668--1674},
  year={2010},
  organization={IEEE}
}

@article{ipopt,
  title={On the implementation of an interior-point filter line-search algorithm for large-scale nonlinear programming},
  author={W{\"a}chter, Andreas and Biegler, Lorenz T},
  journal={Mathematical programming},
  volume={106},
  pages={25--57},
  year={2006},
  publisher={Springer}
}

@article{snopt,
  title={SNOPT: An SQP algorithm for large-scale constrained optimization},
  author={Gill, Philip E and Murray, Walter and Saunders, Michael A},
  journal={SIAM review},
  volume={47},
  number={1},
  pages={99--131},
  year={2005},
  publisher={SIAM}
}

@incollection{knitro,
  title={KNITRO: An integrated package for nonlinear optimization},
  author={Byrd, Richard H and Nocedal, Jorge and Waltz, Richard A},
  booktitle={Large-scale nonlinear optimization},
  pages={35--59},
  year={2006},
  publisher={Springer}
}

@misc{mjx,
  author       = {{MuJoCo XLA Authors}},
  title        = {{MuJoCo XLA (MJX)}},
  howpublished = {\url{https://mujoco.readthedocs.io/en/stable/mjx.html}},
  note         = {Accessed: March 25, 2026},
  year         = {2024}
}

@article{isaac_lab,
  title={Isaac lab: A gpu-accelerated simulation framework for multi-modal robot learning},
  author={Mittal, Mayank and Roth, Pascal and Tigue, James and Richard, Antoine and Zhang, Octi and Du, Peter and Serrano-Munoz, Antonio and Yao, Xinjie and Zurbr{\"u}gg, Ren{\'e} and Rudin, Nikita and others},
  journal={arXiv preprint arXiv:2511.04831},
  year={2025}
}

@article{brax,
  title={Brax--a differentiable physics engine for large scale rigid body simulation},
  author={Freeman, C Daniel and Frey, Erik and Raichuk, Anton and Girgin, Sertan and Mordatch, Igor and Bachem, Olivier},
  journal={arXiv preprint arXiv:2106.13281},
  year={2021}
}

@article{mppi,
  title={Model predictive path integral control: From theory to parallel computation},
  author={Williams, Grady and Aldrich, Andrew and Theodorou, Evangelos A},
  journal={Journal of Guidance, Control, and Dynamics},
  volume={40},
  number={2},
  pages={344--357},
  year={2017},
  publisher={American Institute of Aeronautics and Astronautics}
}

@article{cem,
  title={A tutorial on the cross-entropy method},
  author={De Boer, Pieter-Tjerk and Kroese, Dirk P and Mannor, Shie and Rubinstein, Reuven Y},
  journal={Annals of operations research},
  volume={134},
  number={1},
  pages={19--67},
  year={2005},
  publisher={Springer}
}

@inproceedings{madnlp,
  title={GPU-accelerated dynamic nonlinear optimization with ExaModels and MadNLP},
  author={Pacaud, Fran{\c{c}}ois and Shin, Sungho},
  booktitle={2024 IEEE 63rd Conference on Decision and Control (CDC)},
  pages={5963--5968},
  year={2024},
  organization={IEEE}
}

@article{hykkt,
  title={HyKKT: a hybrid direct-iterative method for solving KKT linear systems},
  author={Regev, Shaked and Chiang, Nai-Yuan and Darve, Eric and Petra, Cosmin G and Saunders, Michael A and {\'S}wirydowicz, Kasia and Pele{\v{s}}, Slaven},
  journal={Optimization Methods and Software},
  volume={38},
  number={2},
  pages={332--355},
  year={2023},
  publisher={Taylor \& Francis}
}

@misc{cudss,
  title={{NVIDIA cuDSS (Preview): A High-Performance CUDA Library for Direct Sparse Solvers}},
  author={{NVIDIA}},
  howpublished={\url{https://docs.nvidia.com/cuda/cudss/index.html}},
  note={Accessed: 2025}
}

@misc{jax,
  author        = "James Bradbury and Roy Frostig and Peter Hawkins and Matthew James Johnson and Chris Leary and Dougal Maclaurin and George Necula and Adam Paszke and Jake Vander{P}las and Skye Wanderman-{M}ilne and Qiao Zhang",
  title         = "{JAX}: composable transformations of {P}ython+{N}um{P}y programs",
  howpublished  = "\url{http://github.com/jax-ml/jax}",
  year          = "2018"
}

@misc{drake,
 author = "Russ Tedrake and the Drake Development Team",
 title = "Drake: Model-based design and verification for robotics",
 year = 2019,
 url = "https://drake.mit.edu"
}

@article{casadi,
  title={CasADi—A software framework for nonlinear optimization and optimal control},
  author={Andersson, Joel and Gillis, Joris and Horn, Greg and Rawlings, Jim and Diehl, Moritz},
  journal={Mathematical Programming Computation},
  volume={11},
  number={1},
  pages={1--36},
  year={2018},
  publisher={Springer Verlag}
}

@article{pyomo, title={Pyomo: modeling and solving mathematical programs in Python}, author={Hart, William E and Watson, Jean-Paul and Woodruff, David L}, journal={Mathematical Programming Computation}, volume={3}, number={3}, pages={219--260}, year={2011}, publisher={Springer} }

@article{jump,
    author = {Miles Lubin and Oscar Dowson and Joaquim {Dias Garcia} and Joey Huchette and Beno{\^i}t Legat and Juan Pablo Vielma},
    title = {{JuMP} 1.0: {R}ecent improvements to a modeling language for mathematical optimization},
    journal = {Mathematical Programming Computation},
    year = {2023},
    doi = {10.1007/s12532-023-00239-3}
}

@article{optnet,
  title={OptNet: Differentiable Optimization as a Layer in Neural Networks},
  author={Brandon Amos and J. Zico Kolter},
  journal={arXiv preprint arXiv:1703.00443},
  year={2017}
}

@misc{moreau,
  author        = "Barratt, Shane and Nobel, Parth and Diamond, Steven",
  title         = "Moreau: {GPU}-Native Differentiable Optimization",
  howpublished  = "\url{https://moreau.so}",
  year          = "2026"
}

@article{mpax,
  title={MPAX: Mathematical Programming in JAX},
  author={Lu, Haihao and Peng, Zedong and Yang, Jinwen},
  journal={arXiv preprint arXiv:2412.09734},
  year={2024}
}

@article{opt2skill,
  title={Opt2skill: Imitating dynamically-feasible whole-body trajectories for versatile humanoid loco-manipulation},
  author={Liu, Fukang and Gu, Zhaoyuan and Cai, Yilin and Zhou, Ziyi and Jung, Hyunyoung and Jang, Jaehwi and Zhao, Shijie and Ha, Sehoon and Chen, Yue and Xu, Danfei and others},
  journal={IEEE Robotics and Automation Letters},
  year={2025},
  publisher={IEEE}
}

@misc{good_old_fashioned_engineering,
  title={Good old-fashioned engineering can close the 100,000-year “data gap” in robotics},
  author={Goldberg, Ken},
  journal={Science Robotics},
  volume={10},
  number={105},
  pages={eaea7390},
  year={2025},
  publisher={American Association for the Advancement of Science}
}

@article{lifted_kkt,
  title={Accelerating optimal power flow with GPUs: SIMD abstraction of nonlinear programs and condensed-space interior-point methods},
  author={Shin, Sungho and Anitescu, Mihai and Pacaud, Fran{\c{c}}ois},
  journal={Electric Power Systems Research},
  volume={236},
  pages={110651},
  year={2024},
  publisher={Elsevier}
}

@article{trust_region_ipm,
  title={A trust region method based on interior point techniques for nonlinear programming},
  author={Byrd, Richard H and Gilbert, Jean Charles and Nocedal, Jorge},
  journal={Mathematical programming},
  volume={89},
  number={1},
  pages={149--185},
  year={2000},
  publisher={Springer}
}

@article{trust_region_pd,
  title={A primal-dual trust-region algorithm for non-convex nonlinear programming},
  author={Conn, Andrew R and Gould, Nicholas IM and Orban, Dominique and Toint, Philippe L},
  journal={Mathematical programming},
  volume={87},
  number={2},
  pages={215--249},
  year={2000},
  publisher={Springer}
}

@book{nonlinear_programming,
  title={Nonlinear programming: sequential unconstrained minimization techniques},
  author={Fiacco, Anthony V and McCormick, Garth P},
  year={1990},
  publisher={SIAM}
}

@article{quadrotor_dynamics,
  title={Quadrotor kinematics and dynamics},
  author={Powers, Caitlin and Mellinger, Daniel and Kumar, Vijay},
  journal={Handbook of unmanned aerial vehicles},
  pages={307--328},
  year={2015},
  publisher={Springer Netherlands}
}

@article{jaxopt,
  title={Efficient and Modular Implicit Differentiation},
  author={Blondel, Mathieu and Berthet, Quentin and Cuturi, Marco and Frostig, Roy
    and Hoyer, Stephan and Llinares-L{\'o}pez, Felipe and Pedregosa, Fabian
    and Vert, Jean-Philippe},
  journal={arXiv preprint arXiv:2105.15183},
  year={2021}
}

@misc{qpax,
      title={A Differentiable Interior-Point Method in Single Precision},
      author={Jon Arrizabalaga and Kevin Tracy and Zachary Manchester},
      year={2026},
      eprint={2605.17913},
      archivePrefix={arXiv},
      primaryClass={math.OC},
      url={https://arxiv.org/abs/2605.17913},
}

@inproceedings{quadruped_mpc_imitate,
  title={Imitation learning from mpc for quadrupedal multi-gait control},
  author={Reske, Alexander and Carius, Jan and Ma, Yuntao and Farshidian, Farbod and Hutter, Marco},
  booktitle={2021 IEEE International Conference on Robotics and Automation (ICRA)},
  pages={5014--5020},
  year={2021},
  organization={IEEE}
}

@article{exatron,
  title={Leveraging GPU batching for scalable nonlinear programming through massive Lagrangian decomposition},
  author={Kim, Youngdae and Pacaud, Fran{\c{c}}ois and Kim, Kibaek and Anitescu, Mihai},
  journal={arXiv preprint arXiv:2106.14995},
  year={2021}
}

@article{pytorch,
  title={Pytorch: An imperative style, high-performance deep learning library},
  author={Paszke, Adam and Gross, Sam and Massa, Francisco and Lerer, Adam and Bradbury, James and Chanan, Gregory and Killeen, Trevor and Lin, Zeming and Gimelshein, Natalia and Antiga, Luca and others},
  journal={Advances in neural information processing systems},
  volume={32},
  year={2019}
}

@article{nlp_survey,
  title={Optimization-based control for dynamic legged robots},
  author={Wensing, Patrick M and Posa, Michael and Hu, Yue and Escande, Adrien and Mansard, Nicolas and Del Prete, Andrea},
  journal={IEEE Transactions on Robotics},
  volume={40},
  pages={43--63},
  year={2023},
  publisher={IEEE}
}

\end{document}